\documentclass[runningheads]{llncs}
\usepackage[T1]{fontenc}
\usepackage{graphicx}
\usepackage{hyperref}
\usepackage{color}

\newcommand\modelname{AViT}
\usepackage{multirow}
\usepackage{amsmath}
\usepackage{amssymb}
\usepackage{booktabs}
\usepackage{breqn}
\usepackage{array}
\newcolumntype{P}[1]{>{\centering\arraybackslash}p{#1}}
\newcolumntype{R}[1]{>{\raggedleft\arraybackslash}p{#1}}
\newcolumntype{L}[1]{>{\raggedright\arraybackslash}p{#1}}
\begin{document}
\title{\modelname{}: Adapting Vision Transformers for Small Skin Lesion Segmentation Datasets}
\titlerunning{\modelname{}}
% \author{Paper ID: 14, Anonymous submission to ISICW 2023}
% %
% \authorrunning{Anonymous et al.}
\authorrunning{S. Du et al.}

\author{Siyi Du\inst{1}\orcidID{0000-0002-9961-4533} \and
Nourhan Bayasi\inst{1}\orcidID{0000-0003-4653-6081} \and
Ghassan Hamarneh\inst{2}\orcidID{0000-0001-5040-7448} \and
Rafeef Garbi\inst{1}\orcidID{0000-0001-6224-0876}}
% \institute{Anonymous Organization \\
% \email{abc@def.hij}}
\institute{University of British Columbia, Vancouver, British Columbia, CA \\
\email{\{siyi,nourhanb,rafeef\}@ece.ubc.ca} \\
\and Simon Fraser University, Burnaby, British Columbia, CA \\
\email{hamarneh@sfu.ca}} 
\maketitle              % typeset the header of the contribution
\begin{abstract}
Skin lesion segmentation (SLS) plays an important role in skin lesion analysis. Vision transformers (ViTs) are considered an auspicious solution for SLS, but they require more training data compared to convolutional neural networks (CNNs) due to their inherent parameter-heavy structure and lack of some inductive biases. To alleviate this issue, current approaches fine-tune pre-trained ViT backbones on SLS datasets, aiming to leverage the knowledge learned from a larger set of natural images to lower the amount of skin training data needed. However, fully fine-tuning all parameters of large backbones is computationally expensive and memory intensive. In this paper, we propose \modelname{}, a novel efficient strategy to mitigate ViTs' data-hunger by transferring any pre-trained ViTs to the SLS task. Specifically, we integrate lightweight modules (adapters) within the transformer layers, which modulate the feature representation of a ViT without updating its pre-trained weights. In addition, we employ a shallow CNN as a prompt generator to create a prompt embedding from the input image, which grasps fine-grained information and CNN's inductive biases to guide the segmentation task on small datasets. Our quantitative experiments on 4 skin lesion datasets demonstrate that \modelname{} achieves competitive, and at times superior, performance to SOTA but with significantly fewer trainable parameters. Our code is available at \url{https://github.com/siyi-wind/AViT}. 
\end{abstract}

% Skin lesion segmentation (SLS) plays an important role in skin lesion analysis. Vision transformers (ViTs) are considered an auspicious solution for SLS, but they require more training data compared to convolutional neural networks (CNNs) due to their inherent parameter-heavy structure and lack of some inductive biases. To alleviate this issue, current approaches fine-tune pre-trained ViT backbones on SLS datasets, aiming to leverage the knowledge learned from a larger set of natural images to lower the amount of skin training data needed. However, fully fine-tuning all parameters of large backbones is computationally expensive and memory intensive. In this paper, we propose \modelname{}, a novel efficient strategy to mitigate ViTs' data-hunger by transferring any pre-trained ViT backbone to the SLS task. Specifically, we integrate lightweight modules (adapters) within the transformer layers, which modulate the feature representation of a ViT without updating its pre-trained weights. In addition, we employ a shallow CNN as a prompt generator to create a prompt embedding from the input image, which grasps fine-grained information and CNN's inductive biases to guide the segmentation task on small datasets. Our quantitative experiments on 4 skin lesion datasets demonstrate that \modelname{} achieves competitive, and at times superior, performance to SOTA but with significantly fewer trainable parameters. Our code will be made available at 
%, especially in the presence of multiple SLS datasets
%
%
\section{Introduction}
Melanoma is the most common and dangerous skin malignancy estimated to cause 97,610 new cases and 7,990 deaths in 2023 the United States alone~\cite{siegel2023cancer}, yet early diagnosis and treatment are highly likely to cure it. Automated skin lesion segmentation (SLS),  which provides thorough qualitative and quantitative information such as location and border, is a challenging and fundamental operation in computer-aided diagnosis~\cite{mirikharaji2023survey}. As a pre-processing step of diagnosis, it boosts the accuracy and robustness of classification by regularizing attention maps~\cite{yan2019melanoma}, offering the region of interest for wide-field images~\cite{birkenfeld2020computer}, or removing lesion-adjacent confounding artifacts~\cite{maron2021reducing,adegun2021deep}. On the other hand, SLS can serve as a simultaneously optimizing task for classification, enabling the models to obtain improved performance on both two tasks~\cite{xie2020mutual}. SLS is also essential for skin color fairness research~\cite{du2022fairdisco}, where the segmented non-lesion area is used to approximate skin tone~\cite{kinyanjui2020fairness}. Vision transformers (ViTs), with their inherent capability to model global image context through the self-attention mechanism, are a set of promising tools to tackle SLS~\cite{gulzar2022skin}. Though ViTs have shown improved performance compared to traditional convolutional neural networks (CNNs)~\cite{li2023transforming}, they are more data-hungry than CNNs, i.e., need more training data, given the lack of some useful inductive biases like weight sharing and locality~\cite{touvron2021training}. This poses a significant challenge in SLS due to the limited availability of training images, where datasets often contain only a few hundred~\cite{mendoncca2013ph} or thousand~\cite{codella2019skin} samples.

% In addition, the localization and delineation of lesions supplied by SLS are essential for both surgical and radiation therapy~\cite{ame2023treat}.
% Melanoma is the most common and dangerous skin malignancy causing around 99,780 new cases and 7,650 deaths in 2022 in the United States alone~\cite{siegel2022cancer}, yet early diagnosis and treatment are highly likely to cure it. Automated skin lesion segmentation (SLS) in dermoscopic images is a fundamental step in computer-aided skin cancer diagnosis and treatment planning~\cite{mirikharaji2022survey,hasan2023survey}. As a pre-processing operation~\cite{adegun2021deep,maron2021reducing} or simultaneously optimizing task of classification~\cite{xie2020mutual}, lesion segmentation enables thorough qualitative and quantitative assessment that can improve accuracy and efficiency of the diagnosis.

To alleviate ViTs’ data-hunger, previous SLS works incorporated some inductive biases through hierarchical architecture~\cite{cao2023swin}, local self-attention \cite{tang2022self}, or convolution layers~\cite{gao2021utnet}. Nevertheless, they trained the models from scratch and overlooked the potential benefits of pre-trained models and valuable information from other domains with abundant data. As transfer learning from ImageNet~\cite{deng2009imagenet} has been demonstrated advantageous for skin lesion tasks~\cite{matsoukas2022makes}, an increasingly popular and promising way is to deploy a large pre-trained ViT as the encoder, and then fine-tune the entire model~\cite{wu2022fat,zhang2021transfuse}. Despite achieving better performance, these techniques that rely on transfer learning have two notable drawbacks. First, a robust ViT typically has plenty of parameters, e.g., ViT-Base (86 million (M))~\cite{dosovitskiy2020image} and Swin-Base (88M)~\cite{liu2021swin}, thus making the full fine-tuning strategy quite expensive in terms of computation and memory requirements, especially when dealing with multiple datasets, i.e., we need to store an entire model for each dataset. Second,  updating all parameters of a large-scale pre-trained model (full fine-tuning) on smaller datasets is found to be unstable~\cite{peters2019tune} and may instead undermine the model's generalizable representations~\cite{yang2023aim}.

% it has been shown
% While been widely used in natural language processing (NLP) for several years, PEFT's application in computer vision has only recently started to gain interest. 
% Jia et al.~\cite{jia2022visual} prepended randomly initialized trainable parameters to the transformer layer's input and .... . 
% Optimizing only those adapters during training, while keeping the rest of the ViT backbone frozen, results in huge computational savings compared to full ViT fine-tuning. 
% This emerging research
The newer parameter-efficient fine-tuning (PEFT) has been proposed as an effective and efficient solution, which only tunes a small subset of the model’s parameters. PEFT in computer vision can be divided into two main directions: 1) prompt tuning~\cite{jia2022visual,bahng2022visual} and 2)  adapter tuning~\cite{yang2023aim,wu2023medical,chen2022adaptformer}. The first direction uses soft (i.e., tunable) prompts: task-specific parameters introduced into the frozen pre-trained ViT backbone's input space and tuned throughout the task-learning process. For example, Jia et al.~\cite{jia2022visual} utilized randomly initialized trainable parameters as soft prompts and prepended them to pre-trained ViT's input for downstream recognition tasks. The second direction uses adapters: trainable lightweight modules inserted into the transformer layers, to modify the hidden representation of the frozen ViT rendering it suitable for a specific task. These PEFT approaches have shown substantially increased efficiency with comparable, or even improved, performance compared to those of full fine-tuning on low-data regimes. Nonetheless, very few works have adapted PEFT to medical imaging. Wu et al.~\cite{wu2023medical} employed adapters to steer the Segment Anything Model (SAM)~\cite{kirillov2023segment}, a promptable ViT-based foundation model trained using 1 billion masks, to medical image segmentation tasks without updating SAM's parameters. However, they require additional pre-training on medical imaging data prior to adaptation as well as hard prompts in the form of un-tunable information input, such as free-form text or a set of foreground/background points, which increases the computational cost and necessitates prior information collection.
% self-supervised 
 % defining the target seeds indicating foreground/background regions
 
To address ViTs' data-hunger while maintaining the model's efficiency, in this work, we propose \modelname{}, a novel transfer learning strategy that adapts a pre-trained ViT backbone to small SLS datasets by using PEFT. We incorporate lightweight adapter modules into the transformer layers to modify the image representation and keep the pre-trained weights untouched. Furthermore, to enhance the information extraction, we introduce a shallow CNN network in parallel with ViT as a prompt generator to generate a prompt embedding from the input image. The prompt captures CNN's valuable inductive biases and fine-grained information, which guides \modelname{} to achieve improved segmentation performance, particularly in scenarios with limited training data. By using ViT-Base as the ViT backbone, the number of tunable parameters of our \modelname{} is 13.6M, which is only 13.7\% of the total \modelname{}'s parameters.

Our contributions can be summarized as follows. (1) To the best of our knowledge, we are the first to introduce PEFT to directly mitigate ViTs' data-hunger in medical image segmentation. (2) We propose \modelname{}, featuring adapters for transferring a pre-trained ViT to the SLS task and a prompt generator for enhancing information extraction. (3) The experimental results on 4 different public datasets indicate that \modelname{} surpasses previous SOTA PEFT algorithms and ViT-based SLS models without pre-trained backbones (gains 2.91\% and 2.32\% on average IOU, respectively). Further, \modelname{} achieves competitive, or even superior performance, to SOTA ViT-based SLS models with pre-trained backbones while having significantly fewer trainable parameters (13.6M vs. 143.5M).
% (raises average IOU by 2.91\%) 

\section{Methodology}
\begin{figure}[t]
\centering
\includegraphics[width=1\linewidth]{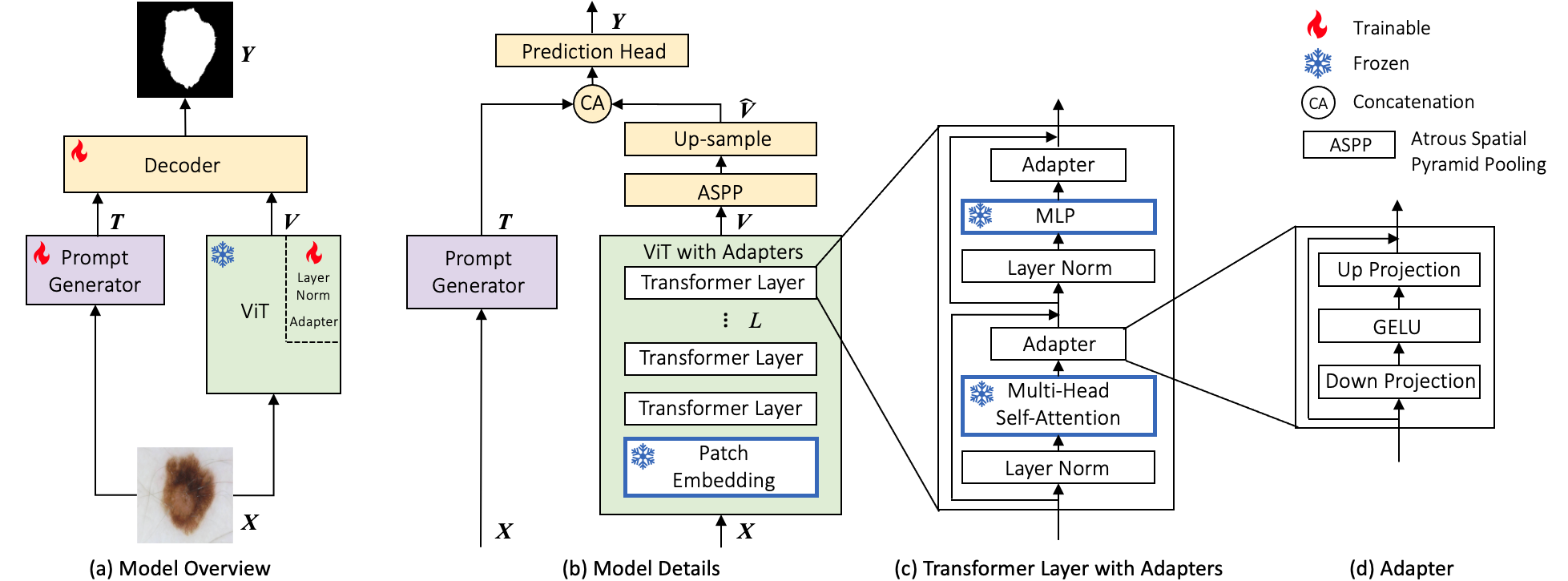}
\caption{Architecture of \modelname{}: (a) Model overview with its prompt generator (a shallow CNN network), a large pre-trained ViT backbone with adapters, and a compact decoder. (b)  Model details. (c) Details of a transformer layer with adapters. (d) Details of our adapters. During training, all modules in (b,c,d) contoured with blue borders are frozen, which encompasses 86.3\% of \modelname{}'s parameters.}
\label{fig:model}
\end{figure}

In skin lesion segmentation (SLS), the model is required to predict a segmentation map $\boldsymbol{Y} \in \{0,1\}^{H \times W}$ that partitions lesion areas based on an RGB skin image $\boldsymbol{X} \in \mathbb{R}^{H \times W \times 3}$. In Fig.~\ref{fig:model}-a, \modelname{} applies a ViT backbone pre-trained on large natural image datasets to the downstream SLS task through adapters and a prompt generator and only optimizes a few parameters. We briefly describe the plain ViT backbone in Section~\ref{sec:pre} and discuss the details of \modelname{} in Section~\ref{sec:model}.

\subsection{Basic ViT}\label{sec:pre}
A plain ViT~\cite{dosovitskiy2020image} backbone contains a patch embedding module and $L$ transformer layers (Fig.~\ref{fig:model}-b). Given an image $\boldsymbol{X}$, the patch embedding module first splits the image into $N$ non-overlapping patches, then flattens and maps them to $D$-dimensional patch embeddings $\boldsymbol{x} \in \mathbb{R}^{N \times D}$ through a linear projection, where $N=\frac{HW}{P^2}$ is the number of patches, and ($P$,$P$) is the patch size. The embedding sequence is then prepended with a learnable [\texttt{class}] token $\boldsymbol{x}_{class}$ to get $\boldsymbol{x}_0=[\boldsymbol{x}_{class};\boldsymbol{x}] \in \mathbb{R}^{(N+1) \times D}$. To utilize the spatial prior, learnable position embeddings $\boldsymbol{E}_{pos} \in \mathbb{R}^{(N+1) \times D}$, defined in~\cite{dosovitskiy2020image}, are added to $\boldsymbol{x}_0$ to get $\boldsymbol{z}_0=\boldsymbol{x}_0+\boldsymbol{E}_{pos}$, which is the input of the first transformer layer. Each transformer layer (Fig.~\ref{fig:model}-c without the adapters) comprises a multi-head self-attention module (MSA) and a multi-layer perceptron module (MLP), along with layer norm (LN). The output of the $l$th transformer layer $\boldsymbol{z}_l \in \mathbb{R}^{(N+1) \times D}$ is: 
\begin{equation}\label{equation:vit}
    \boldsymbol{z}_l'=MSA(LN(\boldsymbol{z}_{l-1}))+\boldsymbol{z}_{l-1} 
\end{equation}
\begin{equation}\label{equation:vit}
    \boldsymbol{z}_l = MLP(LN(\boldsymbol{z}_l'))+\boldsymbol{z}_l'.
\end{equation}
After getting the output of the final transformer layer $\boldsymbol{z}_{L}$, we remove its [\texttt{class}] token and reshape it to a 2D feature representation $\boldsymbol{V} \in \mathbb{R}^{\frac{H}{P} \times \frac{W}{P} \times D}$.
 % and residual connections
% generates a feature representation based on an input image, which c

\subsection{\modelname{}}\label{sec:model}
Given a pre-trained ViT backbone, we integrate adapters in each transformer layer to adjust the generated feature representation $\boldsymbol{V}$ adapted to skin images while leaving the weights of the backbone fixed. In addition, to enhance the information extraction, we employ a prompt generator in parallel, which is a shallow CNN network that produces a prompt embedding $\boldsymbol{T}$ based on the input image. Finally, a lightweight decoder combines $\boldsymbol{V}$ and $\boldsymbol{T}$ to predict a segmentation map. During training, we solely optimize the adapters, prompt generator, layer norm in the ViT backbone, and decoder, which collectively account for 13.7\% of \modelname{}'s parameters. The details of these extensions are as follows.

 % interposed between them
\noindent\textbf{Adapter Tuning:} Similar to~\cite{houlsby2019parameter}, we insert the adapter after MSA and MLP of each transformer layer (Fig.~\ref{fig:model}-c). The adapter (Fig.~\ref{fig:model}-d) contains two linear layers and a GELU function, which first projects the $D$-dimensional input into a smaller dimension $\frac{D}{r}$, where $r$ is the reduction ratio, and projects it back to $D$ dimension, i.e., $Adapter(input)=GELU(input \cdot \boldsymbol{W}_{down})\boldsymbol{W}_{up}$. $ \boldsymbol{W}_{down} \in \mathbb{R}^{D \times \frac{D}{r}}$ and $\boldsymbol{W}_{down} \in \mathbb{R}^{\frac{D}{r} \times D}$. The output of $l$th transformer layer with adapters is:
\begin{equation}\label{equation:vit}
\boldsymbol{z}_l'=Adapter(MSA(LN(\boldsymbol{z}_{l-1})))+\boldsymbol{z}_{l-1} 
\end{equation}
\begin{equation}\label{equation:vit}
    \boldsymbol{z}_l = Adapter(MLP(LN(\boldsymbol{z}_l')))+\boldsymbol{z}_l'.
\end{equation}

\noindent\textbf{Information Enhancement by Prompt Tuning:} Inspired by the prompt tuning~\cite{jia2022visual,gao2022visual}, we deploy soft prompts to extract more information from images and enrich the SLS task learning. Specifically, we utilize the first stage of a ResNet-34 (including 7 convolutional layers) as the prompt generator to automatically create a prompt embedding $\boldsymbol{T}$ from the input image. The prompt is hypothesized to grasp CNN's helpful inductive biases and fine-grained information, e.g., spatial details, boundaries, and texture, to facilitate \modelname{}'s segmentation ability despite the small training datasets. Our soft prompt produced by the network is more flexible, customized to each input image, and includes rich information, in contrast to previous soft prompts that are simple free tunable parameters and remain constant for all inputs. Moreover, it is worth noting that our prompt generator has only a small number of parameters (0.23M), which is different from previous hybrid models combining a ViT with a large CNN backbone, e.g., ResNet-34 (21.3M)~\cite{wu2022fat,he2023h2former} or ResNet-50 (23.5M)~\cite{wang2021boundary}.

\noindent\textbf{Lightweight Decoder:} We incorporate a compact decoder for efficient prediction, as opposed to prior works that use complex decoding architectures involving multi-stage up-sampling, convolutional operations, and skip connections~\cite{hatamizadeh2022unetr,wu2022fat}. This choice is driven by the powerful and over-parameterized nature of large pre-trained ViT backbones, which have demonstrated strong transferability to downstream tasks~\cite{matsoukas2022makes}. As visualized in Fig.~\ref{fig:model}-b, after getting the feature representation $\boldsymbol{V}$ from the ViT backbone and the prompt embedding $\boldsymbol{T}$ from the prompt generator, we first pass $\boldsymbol{V}$ through the atrous spatial pyramid pooling module (ASPP) proposed in~\cite{chen2017deeplab}, which uses multiple parallel dilated convolutional layers with different dilation rates, to obtain a feature that extracts local information while capturing lesion context at different scales. After that, we up-sample the output feature of ASPP to get $\hat{\boldsymbol{V}}$, which has the same resolution as $\boldsymbol{T}$. Finally, $\hat{\boldsymbol{V}}$ is concatenated with $\boldsymbol{T}$ and sent to a projection head, which is formed by 3 convolutional layers connected by ReLU activation functions.

% As visualized in Fig.~\ref{fig:model}-b, after obtaining the feature representation $\boldsymbol{V}$ from the ViT backbone and the prompt embedding $\boldsymbol{T}$ from the prompt generator, we first pass $\boldsymbol{V}$ through the atrous spatial pyramid pooling module (ASPP) proposed in~\cite{chen2017deeplab}, which uses multiple parallel dilated convolutional layers with different dilation rates, to obtain $\hat{\boldsymbol{V}}$ that extracts local information while capturing lesion context at different scales. After that, we up-sample the output feature of ASPP to the same resolution as $\boldsymbol{T}$. The newly acquired representation $\hat{\boldsymbol{V}}$ is concatenated with $\boldsymbol{T}$ and sent to a projection head, which is formed by 3 convolutional layers connected by ReLU activation functions.

\section{Experiments}
\noindent\textbf{Datasets and Evaluation Metrics:} We evaluate our \modelname{} on 4 public SLS databases collected from different sources: ISIC 2018 (ISIC)~\cite{codella2019skin}, Dermofit Image Library (DMF)~\cite{ballerini2013color}, Skin Cancer Detection (SCD)~\cite{glaister2013msim}, and PH2~\cite{mendoncca2013ph}, which contain 2594, 1212, 206, and 200 skin images along with their segmentation maps, respectively. We perform 5-fold cross-validation and measure our model's segmentation performance using Dice and IOU metrics, computational cost at inference via gigaFLOPs (GFLOPs), and memory footprint via the number of needed parameters. Due to the table width restriction and the high number of columns, we only report the standard deviation (std) of average Dice and IOU in tables and provide the std for each dataset in the supplementary material.

\noindent\textbf{Implementation Details:} We resize the images to $224 \times 224$ and augment them by random scaling, shifting, rotation, flipping, Gaussian noise, and brightness and contrast changes. The ViT backbone of \modelname{} is a ViT-B/16~\cite{dosovitskiy2020image}, with a patch size of $16 \times 16$, pre-trained on ImageNet-21k. Similar to~\cite{yang2023aim}, the reduction ratio $r$ of the adapters is 4. The output dimension of ASPP is 256. All models are deployed on a single TITAN V and trained using a combination of Dice and binary cross entropy loss~\cite{du2023mdvit,taghanaki2019combo} for 200 epochs with the AdamW optimizer~\cite{loshchilov2017decoupled}, a batch size of 16, and an initial learning rate of $1 \times 10^{-4}$, which changes through a linear decay scheduler whose step size is 50 and decay factor $\gamma=0.5$.

\begin{table*}[t]
\centering
\caption{Skin lesion segmentation (SLS) results comparing BASE (\modelname{} w/o both adapters and the prompt generator and is fully fine-tuned), \modelname{}, and SOTA algorithms. We report the models' parameter count in millions (M). The 2nd column shows which pre-trained backbone the model used. R-34/50 represents ResNet-34/50.}
\label{table:SOTA}
\resizebox{\textwidth}{!}{
\begin{tabular}{|p{20mm}|P{21mm}|P{18mm}R{13.5mm}P{11mm}|P{10.5mm}P{10.5mm}P{10.5mm}P{10.5mm}P{13mm}|P{10.5mm}P{10.5mm}P{10.5mm}P{10.5mm}P{13mm}|}
\hline 
\textbf{Model} & \textbf{Pre-}  & \multicolumn{1}{c}{\textbf{\#Total}} & \multicolumn{1}{c}{\textbf{\#Tuned}} & \textbf{GFL-} & \multicolumn{10}{c|}{\textbf{Segmentation Results in Test Sets (\%)}} \\
\cline{6-15}
~ & \textbf{trained} & \multicolumn{1}{c}{\textbf{Param.}} & \multicolumn{1}{c}{\textbf{Param.}} & \textbf{OPs} & \multicolumn{5}{c|}{\textbf{Dice $\uparrow$}} & \multicolumn{5}{c|}{\textbf{IOU $\uparrow$}} \\
\cline{6-15}
~ & \textbf{backbone} & \multicolumn{1}{c}{\textbf{(M) $\downarrow$}} & \multicolumn{1}{c}{\textbf{(M) $\downarrow$}} & \textbf{$\downarrow$} &  ISIC & DMF & SCD & PH2 & Avg\tiny{$\pm$\emph{std}}  & ISIC & DMF & SCD & PH2 & Avg\tiny{$\pm$\emph{std}}\\
\hline 
\multicolumn{15}{|c|}{\textbf{(a) Full Fine-tuned BASE \& Proposed PEFT Method}} \\ \hline
BASE	&	ViT-B	&	91.8$\times$	&	91.8$\times$	&	18.0 	&	90.77 	&	91.69 	&	91.95 	&	95.64 	&	92.51\tiny{\emph{0.22}} 	&	83.71 	&	84.89 	&	85.42 	&	91.72 	&	86.43\tiny{\emph{0.34}} 	\\
\modelname{}	&	ViT-B	&99.4 (13.6$\times$)&	13.6$\times$	&	20.9 	&	\underline{91.74} 	&	\underline{92.04} 	&	93.16 	&	95.66 	&	93.15\tiny{\emph{0.42}} 	&	\underline{85.22} 	&	\underline{85.47} 	&	87.39 	&	91.72 	&	87.45\tiny{\emph{0.70}} 	\\ \hline
\multicolumn{15}{|c|}{\textbf{(b) PEFT Methods}} \\ \hline
VPT	&	ViT-B	&	92.8 (7.0$\times$)	&	7.0$\times$	&	26.5 	&	90.89 	&	91.26 	&	89.09 	&	93.14 	&	91.10\tiny{\emph{0.46}} 	&	83.83 	&	84.14 	&	80.76 	&	87.27 	&	84.00\tiny{\emph{0.74}} 	\\
AdaptFormer	&	ViT-B	&	93.0 (7.2$\times$)	&	7.2$\times$	&	18.2 	&	91.12 	&	91.27 	&	89.65 	&	93.76 	&	91.45\tiny{\emph{0.42}} 	&	84.15 	&	84.18 	&	81.49 	&	88.33 	&	84.54\tiny{\emph{0.67}} 	\\ \hline

\multicolumn{15}{|c|}{\textbf{(c) SLS Methods w/o Pre-trained Backbones \& Trained From Scratch}} \\ \hline
SwinUnet	&	None	&	41.4$\times$	&	41.4$\times$	&	8.7 	&	89.64 	&	90.67 	&	89.77 	&	94.24 	&	91.08\tiny{\emph{0.57}} 	&	81.94 	&	83.19 	&	82.07	&	89.24 	&	84.11\tiny{\emph{0.79}} 	\\
UNETR  &  None  &  87.7$\times$  & 87.7$\times$  &  20.2  & 89.60 	&	90.53 	&	88.13 	&	93.92 	&	90.55\tiny{\emph{0.87}} 	&	81.86 	&	83.02 	&	79.96 	&	88.68 	&	83.38\tiny{\emph{1.24}} 	  \\ 
UTNet	&	None	&	10.0$\times$	&	10.0$\times$	&	13.2 	&	89.68 	&	89.87 	&	88.11 	&	93.29 	&	90.23\tiny{\emph{0.61}} 	&	81.99 	&	81.91 	&	79.71 	&	87.62 	&	82.81\tiny{\emph{0.77}} 	\\
MedFormer	&	None	&	19.2$\times$	&	19.2$\times$	&	13.0 	&	90.47 	&	90.85 	&	90.60 	&	94.82 	&	91.68\tiny{\emph{0.74}} 	&	83.22 	&	83.52 	&	83.53 	&	90.23 	&	85.13\tiny{\emph{1.12}} 	\\
Swin UNETR	&	None	&	25.1$\times$	&	25.1$\times$	&	14.3 	&	90.19 	&	91.00 	&	90.71 	&	94.54 	&	91.61\tiny{\emph{0.49}} 	&	82.78 	&	83.77 	&	83.54 	&	89.74 	&	84.96\tiny{\emph{0.74}} 	\\ \hline

\multicolumn{15}{|c|}{\textbf{(d) SLS Methods w/ Pre-trained Backbones \& Fully Fine-tuned}} \\ \hline
H2Former	&	R-34	&	33.7$\times$	&	33.7$\times$	&	24.7 	&	91.17 	&	91.29 	&	92.76 	&	95.65 	&	92.72\tiny{\emph{0.63}} 	&	84.35 	&	84.22 	&	87.04 	&	91.77 	&	86.85\tiny{\emph{0.91}} 	\\
FAT-Net	&	R-34, DeiT-T	&	28.8$\times$	&	28.8$\times$	&	42.8 	&	91.26 	&	91.32 	&	93.03 	&	96.07 	&	92.92\tiny{\emph{0.48}} 	&	84.42 	&	84.25 	&	87.23 	&	92.48 	&	87.10\tiny{\emph{0.80}} 	\\
BAT	&	R-50	&	46.2$\times$	&	46.2$\times$	&	10.3 	&	91.33 	&	91.20 	&	92.95 	&	95.84 	&	92.83\tiny{\emph{0.46}} 	&	84.40 	&	84.03 	&	87.08 	&	92.04 	&	86.89\tiny{\emph{0.78}} 	\\
TransFuse	&	R-50, DeiT-B	&	143.5$\times$	&	143.5$\times$	&	63.4 	&	91.73 	&	91.96 	&	\underline{94.11} 	&	\underline{96.18} 	&	\underline{93.50}\tiny{\emph{0.27}} 	&	\underline{85.22} 	&	85.33 	&	\underline{89.03} 	&	\underline{92.69} 	&	\underline{88.07}\tiny{\emph{0.47}} 	\\ \hline
\end{tabular}}
\end{table*}

\noindent\textbf{Comparing Against The Baseline (BASE):} BASE is established by removing the adapters and the prompt generator of \modelname{} and optimizing all the parameters during training. In Table~\ref{table:SOTA}-a, \modelname{} achieves superior performance compared to BASE, with average IOU and Dice improvements of 1.02\% and 0.64\%, respectively, while utilizing significantly fewer trainable parameters (13.6M vs. 91.8M). This suggests that BASE exhibits overfitting, and full fine-tuning is unsuitable for transferring knowledge to smaller skin datasets, whereas \modelname{} effectively leverages the learnt knowledge and demonstrates strong generalization capability on the SLS task. When considering the memory requirements for the 4 datasets, BASE would require storing 4 entirely new models, resulting in a total of $91.8 \times 4=367.2$M parameters. On the contrary, \modelname{} only needs to store the pre-trained ViT backbone once, resulting in reduced storage needs, i.e., $85.8+13.6\times4=140.2$M. As the number of domains increases, the memory savings offered by \modelname{} compared to BASE will become even more pronounced.

\begin{figure}[t]
\centering
\includegraphics[width=1\linewidth]{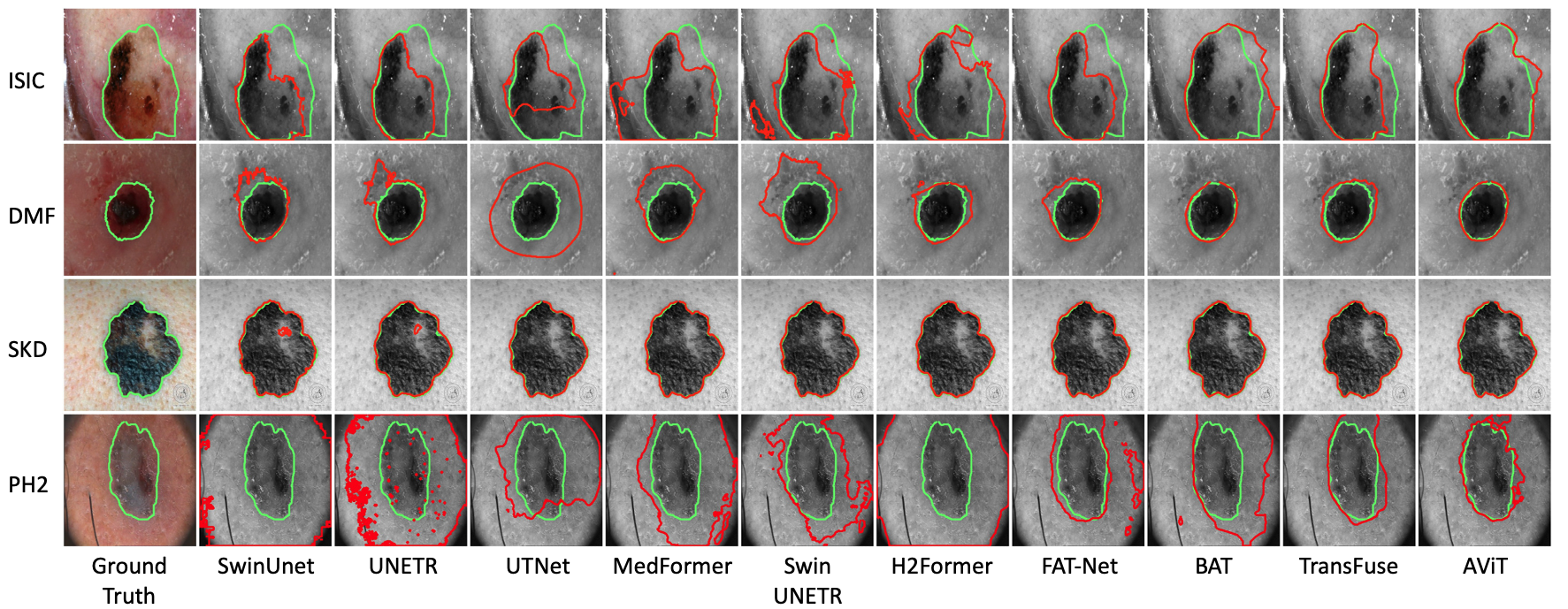}
\caption{Visual comparison with different SOTA methods. The green contours are the ground truth, and the red contours are the segmentation results.}
\label{fig:visualization}
\end{figure}

\noindent\textbf{Comparing Against State-of-the-Art (SOTA) Methods:} We conduct experiments on SOTA PEFT and SLS approaches. We first reproduced VPT~\cite{jia2022visual} that added learnable visual prompts in the input space and AdaptFormer~\cite{chen2022adaptformer} that introduced adapters in the transformer layers. We set the number of prompts in VPT to 100. In Table~\ref{table:SOTA}-b, \modelname{} surpasses them across all datasets (gains 2.91\% on average IOU over AdaptFormer), with comparable trainable parameters. 

Additionally, we compare various ViT-based SLS algorithms and divide them into two groups. \emph{Group 1} is models without pre-trained backbones and trained from scratch:  SwinUnet~\cite{cao2023swin}, UNETR~\cite{hatamizadeh2022unetr}, UTNet~\cite{gao2021utnet}, MedFormer~\cite{gao2022data}, and Swin UNETR~\cite{tang2022self}. \emph{Group 2} is models with pre-trained backbones and fully fine-tuned: H2Former~\cite{he2023h2former}, FAT-Net~\cite{wu2022fat}, BAT~\cite{wang2021boundary}, and TransFuse~\cite{zhang2021transfuse}. H2Former and BAT used pre-trained ResNet but randomly initialized transformer modules. Table~\ref{table:SOTA}-c shows that \modelname{} outperforms \emph{Group 1} across all datasets by a large margin (increases average IOU of MedFormer by 2.32\%), with comparable and even fewer trainable parameters (13.6M vs. 19.2M). Table~\ref{table:SOTA}-d illustrates that \modelname{} achieves competitive or higher segmentation performance compared to \emph{Group 2}, with fewer trainable parameters. For instance, \modelname{} achieves a marginally lower average Dice compared to TransFuse (0.35\% difference), yet its parameter count and computational complexity (GFLOPs) are 1/10 and 1/3 less than that of TransFuse, respectively. Fig.~\ref{fig:visualization}
visualizes \modelname{}'s segmentation performance.

\begin{table*}[t]
\centering
\caption{Experiments using different pre-trained ViT backbones and ablation study of \modelname{}. $^*$ means the pre-trained backbone is frozen throughout training. $^{-P}$ or $^{-A}$ represent not using the prompt generator or adapters in \modelname{}.}
\label{table:ablation}
\resizebox{\textwidth}{!}{
\begin{tabular}{|p{14mm}|P{15mm}|P{20mm}R{13.5mm}P{11mm}|P{10.5mm}P{10.5mm}P{10.5mm}P{10.5mm}P{13mm}|P{10.5mm}P{10.5mm}P{10.5mm}P{10.5mm}P{13mm}|}
\hline 
\textbf{Model} &  \textbf{Pre-} & \multicolumn{1}{c}{\textbf{\#Total}} & \multicolumn{1}{c}{\textbf{\#Tuned}} & \textbf{GFL-}  & \multicolumn{10}{c|}{\textbf{Segmentation Results in Test Sets (\%)}} \\
\cline{6-15}
~ & \textbf{trained} &  \multicolumn{1}{c}{\textbf{Param.}} & \multicolumn{1}{c}{\textbf{Param.}} & \textbf{OPs} & \multicolumn{5}{c|}{\textbf{Dice $\uparrow$}} & \multicolumn{5}{c|}{\textbf{IOU $\uparrow$}} \\
\cline{6-15}
~ & \textbf{backbone} & \multicolumn{1}{c}{\textbf{(M)}}  & \multicolumn{1}{c}{\textbf{(M)}} & ~ &  ISIC & DMF & SCD & PH2 & Avg\tiny{$\pm$\emph{std}}  & ISIC & DMF & SCD & PH2 & Avg\tiny{$\pm$\emph{std}}\\
\hline
\multicolumn{15}{|c|}{\textbf{(a) Applicability to Various Pre-trained ViT Backbones}} \\ \hline
BASE	&	Swin-B	&	63.8$\times$	&	63.8$\times$	&	15.6 	&	91.63 	&	91.70 	&	92.71 	&	95.88 	&	92.98\tiny{\emph{0.37}} 	&	85.05 	&	84.89 	&	86.60 	&	92.13 	&	87.17\tiny{\emph{0.61}}  	\\
\modelname{}	&	Swin-B	&	68.9 (9.5$\times$)	&	9.5$\times$	&	18.3 	&	91.54 	&	91.73 	&	93.60 	&	95.68 	&	93.14\tiny{\emph{0.39}}  	&	84.90 	&	84.94 	&	88.12 	&	91.77 	&	87.43\tiny{\emph{0.64}}  	\\
BASE	&	Swin-L	&139.8$\times$	&	139.8$\times$&	32.3 	&	91.64 	&	91.69 	&	92.93 	&	95.83 	&	93.02\tiny{\emph{0.25}}  	&	85.08 	&	84.86 	&	86.97 	&	92.04 	&	87.24\tiny{\emph{0.43}} 	\\
\modelname{}	&	Swin-L	&151.1 (17.6$\times$)	&	17.6$\times$	&	36.3 	&	91.56 	&	91.91 	&	93.74 	&	96.07 	&	93.32\tiny{\emph{0.31}}  	&	84.93 	&	85.24 	&	88.38 	&	92.47 	&	87.76\tiny{\emph{0.50}}  	\\
BASE	&	ViT-L	&	311.2$\times$	&	311.2$\times$	&	61.2 	&	91.37 	&	91.76 	&	93.23 	&	95.86 	&	93.06\tiny{\emph{0.29}}  	&	84.60 	&	84.99 	&	87.52 	&	92.09 	&	87.30\tiny{\emph{0.47}}  	\\
\modelname{}	&	ViT-L	&	336.9 (33.7$\times$)	&	33.7$\times$	&	67.7 	&	91.54 	&	91.77 	&	93.48 	&	95.73 	&	93.13\tiny{\emph{0.48}}  	&	84.88 	&	85.01 	&	87.94 	&	91.85 	&	87.42\tiny{\emph{0.79}}  	\\  
BASE	&	DeiT-B	&	91.8$\times$	&	91.8$\times$	&	18.0 	&	91.48 	&	91.82 	&	93.63 	&	95.83 	&	92.94\tiny{\emph{0.32}}  	&	84.77 	&	85.10 	&	86.53 	&	92.04 	&	87.11\tiny{\emph{0.52}}  	\\
\modelname{}	&	DeiT-B	&	99.4 (13.6$\times$)	&	13.6$\times$	&	20.9 	&	91.70 	&	91.85 	&	93.67 	&	95.97 	&	93.30\tiny{\emph{0.31}}  	&	85.14 	&	85.17 	&	88.22 	&	92.30 	&	87.71\tiny{\emph{0.51}}  	\\  \hline
\multicolumn{15}{|c|}{\textbf{(b) Ablation Study}} \\ \hline
BASE$^*$	&	ViT-B	&	91.8 (6.0$\times$)	&	6.0$\times$	&	18.0 	&	87.18 	&	89.23 	&	86.24 	&	90.17 	&	88.20\tiny{\emph{0.46}}  	&	77.92 	&	80.81 	&	76.27 	&	82.30 	&	79.33\tiny{\emph{0.65}}  	\\
\modelname{}$^{-P}$	&	ViT-B	&	98.9 (13.2$\times$)	&	13.2$\times$	&	19.4 	&	91.47 	&	91.80 	&	91.18 	&	94.75 	&	92.30\tiny{\emph{0.31}}  	&	84.74 	&	85.04 	&	83.98 	&	90.09 	&	85.96\tiny{\emph{0.48}}  	\\
\modelname{}$^{-A}$	&	ViT-B	&	92.3 (6.5$\times$)	&	6.5$\times$	&	19.5 	&	90.87 	&	91.00 	&	89.09 	&	93.87 	&	91.21\tiny{\emph{0.83}}  	&	83.78 	&	83.72 	&	81.18 	&	88.53 	&	84.30\tiny{\emph{1.19}}  	\\
\modelname{}	&	ViT-B	&	99.4 (13.6$\times$)	&	13.6$\times$	&	20.9 	&	\underline{91.74} 	&	\underline{92.04} 	&	\underline{93.16} 	&	\underline{95.66} 	&	\underline{93.15}\tiny{\emph{0.42}} 	&	\underline{85.22} 	&	\underline{85.47} 	&	\underline{87.39} 	&	\underline{91.72} 	&	\underline{87.45}\tiny{\emph{0.70}} 	\\ \hline
\end{tabular}}
\end{table*}

\noindent\textbf{\modelname{} on Different Pre-trained ViT Backbones:} We conduct experiments using ViTs in varied sizes, structures, or training strategies, including ViT-L/16, Swin-B, Swin-L~\cite{liu2021swin}, and DeiT-B, as the pre-trained backbone. For Swin-B/L, we use the output of its 3rd stage as the encoded image feature, whose resolution is the same as ViT-B's output feature. DeiT-B and ViT-B have the same architecture but different training strategies. In Table~\ref{table:ablation}-a, for each ViT backbone, \modelname{} achieves competitive and even higher performance compared to fully fine-tuned BASE, but with substantially fewer parameters (trainable and total) for the 4 datasets, indicating the applicability of our method on different ViTs.

\noindent\textbf{Ablation Study:} To show the efficacy of our proposed components in Section~\ref{sec:model}, we freeze the parameters of BASE's pre-trained ViT to get BASE$^*$ and remove the adapters and prompt generator in \modelname{} to get \modelname{}$^{-A}$ and \modelname{}$^{-P}$, respectively. In Table~\ref{table:ablation}-b, BASE$^*$ attains average Dice and IOU of 88.20\% and 79.33\%, respectively. However, it still falls far behind fully fine-tuned BASE with 92.51\% and 86.43\% on average Dice and IOU, respectively. After adding adapters to BASE (\modelname{}$^{-P}$), the average Dice and IOU increase by 4.10\% and 6.63\%, respectively; after adding a prompt generator to BASE (\modelname{}$^{-A}$), the average Dice and IOU increase by 3.01\% and 4.97\%, respectively. Finally, \modelname{} achieves the highest segmentation results and significantly outperforms BASE$^*$ (increases average Dice and IOU by 4.95\% and 8.12\%, respectively) with only 7.6M more trainable parameters. The above results reveal that our proposed mechanisms boost the segmentation performance, and a combination of both performs best.

% BASE$^*$ attains average Dice and IOU of 88.20\% and 79.33\%, respectively. However, it still falls far behind fully fine-tuned BASE, which achieves 92.51\% and 86.43\% on average Dice and IOU, respectively. 

% After adding a prompt generator to BASE (\modelname{}$^{-A}$), the average Dice and IOU increase by 3.01\% and 4.97\%, respectively; after adding adapters to BASE (\modelname{}$^{-P}$), the average Dice and IOU increase by 4.10\% and 6.63\%, respectively.

\section{Conclusion}
We propose \modelname{}, a new method to alleviate ViTs' data-hunger and apply it on small skin lesion segmentation (SLS) datasets by employing a pre-trained ViT backbone whilst keeping computation and storage memory costs very low via parameter-efficient fine-tuning (PEFT). Specifically, we integrate adapters into the transformer layers to modulate the backbone's image representation without updating its pre-trained weights and utilize a prompt generator to produce a prompt embedding, which captures CNNs' inductive biases and fine-grained information to guide \modelname{} for segmenting skin images on limited data. Our experiments on 4 datasets illustrate that \modelname{} outperforms other PEFT methods and achieves comparable or even superior performance to SOTA SLS approaches but with considerably fewer trainable and total parameters. Moreover, the experiments using different ViT backbones and an ablation study showcase the applicability of \modelname{} and the effectiveness of \modelname{}'s components. Future work will focus on improving \modelname{}'s architecture so that it can achieve SOTA segmentation performance while retaining computation and memory efficiency.

\bibliographystyle{splncs04}
\bibliography{mybibliography}

\begin{thebibliography}{10}
\providecommand{\url}[1]{\texttt{#1}}
\providecommand{\urlprefix}{URL }
\providecommand{\doi}[1]{https://doi.org/#1}

\bibitem{adegun2021deep}
Adegun, A., Viriri, S.: Deep learning techniques for skin lesion analysis and
  melanoma cancer detection: a survey of state-of-the-art. Artificial
  Intelligence Review  \textbf{54},  811--841 (2021)

\bibitem{bahng2022visual}
Bahng, H., Jahanian, A., et~al.: Visual prompting: Modifying pixel space to
  adapt pre-trained models. arXiv preprint arXiv:2203.17274  (2022)

\bibitem{ballerini2013color}
Ballerini, L., Fisher, R.B., Aldridge, B., Rees, J.: A color and texture based
  hierarchical {K-NN} approach to the classification of non-melanoma skin
  lesions. In: Color medical image analysis, pp. 63--86. Springer (2013)

\bibitem{birkenfeld2020computer}
Birkenfeld, J.S., Tucker-Schwartz, J.M., et~al.: Computer-aided classification
  of suspicious pigmented lesions using wide-field images. Computer methods and
  programs in biomedicine  \textbf{195},  105631 (2020)

\bibitem{cao2023swin}
Cao, H., Wang, Y., et~al.: {SwinUnet}: Unet-like pure transformer for medical
  image segmentation. In: ECCV 2022 Workshops. pp. 205--218. Springer (2023)

\bibitem{chen2017deeplab}
Chen, L.C., Papandreou, G., et~al.: {DeepLab}: Semantic image segmentation with
  deep convolutional nets, atrous convolution, and fully connected {CRFs}. IEEE
  transactions on pattern analysis and machine intelligence  \textbf{40}(4),
  834--848 (2017)

\bibitem{chen2022adaptformer}
Chen, S., Ge, C., Tong, Z., Wang, J., Song, Y., et~al.: {AdaptFormer}: Adapting
  vision transformers for scalable visual recognition. In: NeurIPS 2022 (2022)

\bibitem{codella2019skin}
Codella, N., Rotemberg, V., Tschandl, P., Celebi, M.E., Dusza, S., et~al.: Skin
  lesion analysis toward melanoma detection 2018: A challenge hosted by the
  international skin imaging collaboration ({ISIC}). arXiv preprint
  arXiv:1902.03368  (2019)

\bibitem{deng2009imagenet}
Deng, J., Dong, W., Socher, R., Li, L.J., Li, K., Fei-Fei, L.: {ImageNet}: A
  large-scale hierarchical image database. In: CVPR 2009. pp. 248--255. Ieee
  (2009)

\bibitem{dosovitskiy2020image}
Dosovitskiy, A., Beyer, L., Kolesnikov, A., et~al.: An image is worth 16x16
  words: Transformers for image recognition at scale. In: ICLR 2020 (2020)

\bibitem{du2023mdvit}
Du, S., Bayasi, N., Harmarneh, G., Garbi, R.: {MDViT}: Multi-domain vision
  transformer for small medical image segmentation datasets. arXiv preprint
  arXiv:2307.02100  (2023)

\bibitem{du2022fairdisco}
Du, S., Hers, B., Bayasi, N., et~al.: {FairDisCo}: Fairer {AI} in dermatology
  via disentanglement contrastive learning. In: ECCVW 2022. pp. 185--202.
  Springer (2022)

\bibitem{gao2022visual}
Gao, Y., Shi, X., Zhu, Y., Wang, H., et~al.: Visual prompt tuning for test-time
  domain adaptation. arXiv preprint arXiv:2210.04831  (2022)

\bibitem{gao2021utnet}
Gao, Y., Zhou, M., Metaxas, D.N.: {UTNet}: a hybrid transformer architecture
  for medical image segmentation. In: MICCAI 2021. pp. 61--71. Springer (2021)

\bibitem{gao2022data}
Gao, Y., et~al.: A data-scalable transformer for medical image segmentation:
  architecture, model efficiency, and benchmark. arXiv preprint
  arXiv:2203.00131  (2022)

\bibitem{glaister2013msim}
Glaister, J., Amelard, R., Wong, A., Clausi, D.A.: {MSIM}: Multistage
  illumination modeling of dermatological photographs for
  illumination-corrected skin lesion analysis. IEEE Transactions on Biomedical
  Engineering  \textbf{60}(7),  1873--1883 (2013)

\bibitem{gulzar2022skin}
Gulzar, Y., Khan, S.A.: Skin lesion segmentation based on vision transformers
  and convolutional neural networks—a comparative study. Applied Sciences
  \textbf{12}(12), ~5990 (2022)

\bibitem{hatamizadeh2022unetr}
Hatamizadeh, A., Tang, Y., Nath, V., Yang, D., et~al.: {UNETR}: Transformers
  for {3D} medical image segmentation. In: WACV 2022. pp. 574--584 (2022)

\bibitem{he2023h2former}
He, A., Wang, K., et~al.: {H2Former}: An efficient hierarchical hybrid
  transformer for medical image segmentation. IEEE Transactions on Medical
  Imaging  (2023)

\bibitem{houlsby2019parameter}
Houlsby, N., Giurgiu, A., Jastrzebski, S., Morrone, B., et~al.:
  Parameter-efficient transfer learning for {NLP}. In: ICML 2019. pp.
  2790--2799. PMLR (2019)

\bibitem{jia2022visual}
Jia, M., Tang, L., Chen, B.C., Cardie, C., Belongie, S., Hariharan, B., Lim,
  S.N.: Visual prompt tuning. In: ECCV 2022. pp. 709--727. Springer (2022)

\bibitem{kinyanjui2020fairness}
Kinyanjui, N.M., Odonga, T., et~al.: Fairness of classifiers across skin tones
  in dermatology. In: MICCAI 2020. pp. 320--329. Springer (2020)

\bibitem{kirillov2023segment}
Kirillov, A., et~al.: Segment anything. arXiv preprint arXiv:2304.02643  (2023)

\bibitem{li2023transforming}
Li, J., Chen, J., Tang, Y., Wang, C., Landman, B.A., Zhou, S.K.: Transforming
  medical imaging with transformers? a comparative review of key properties,
  current progresses, and future perspectives. Medical image analysis p. 102762
  (2023)

\bibitem{liu2021swin}
Liu, Z., Lin, Y., Cao, Y., Hu, H., et~al.: {Swin Transformer}: Hierarchical
  vision transformer using shifted windows. In: ICCV 2021. pp. 10012--10022
  (2021)

\bibitem{loshchilov2017decoupled}
Loshchilov, I., Hutter, F.: Decoupled weight decay regularization. arXiv
  preprint arXiv:1711.05101  (2017)

\bibitem{maron2021reducing}
Maron, R.C., Hekler, A., Krieghoff-Henning, E., Schmitt, M., et~al.: Reducing
  the impact of confounding factors on skin cancer classification via image
  segmentation: technical model study. Journal of Medical Internet Research
  \textbf{23}(3),  e21695 (2021)

\bibitem{matsoukas2022makes}
Matsoukas, C., Haslum, J.F., et~al.: What makes transfer learning work for
  medical images: feature reuse \& other factors. In: CVPR 2022. pp. 9225--9234
  (2022)

\bibitem{mendoncca2013ph}
Mendon{\c{c}}a, T., Ferreira, P.M., et~al.: {PH 2-A} dermoscopic image database
  for research and benchmarking. In: EMBC 2013. pp. 5437--5440. IEEE (2013)

\bibitem{mirikharaji2023survey}
Mirikharaji, Z., Abhishek, K., Bissoto, A., Barata, C., et~al.: A survey on
  deep learning for skin lesion segmentation. Medical Image Analysis
  \textbf{88},  102863 (2023)

\bibitem{peters2019tune}
Peters, M.E., Ruder, S., Smith, N.A.: To tune or not to tune? adapting
  pretrained representations to diverse tasks. ACL 2019 p.~7 (2019)

\bibitem{siegel2023cancer}
Siegel, R.L., Miller, K.D., Wagle, N.S., Jemal, A.: Cancer statistics, 2023.
  CA: a cancer journal for clinicians  \textbf{73}(1),  17--48 (2023)

\bibitem{taghanaki2019combo}
Taghanaki, S.A., Zheng, Y., Zhou, S.K., Georgescu, B., Sharma, P., Xu, D.,
  et~al.: Combo loss: Handling input and output imbalance in multi-organ
  segmentation. Computerized Medical Imaging and Graphics  \textbf{75},  24--33
  (2019)

\bibitem{tang2022self}
Tang, Y., Yang, D., Li, W., Roth, H.R., Landman, B., Xu, D., Nath, V.,
  Hatamizadeh, A.: Self-supervised pre-training of swin transformers for {3D}
  medical image analysis. In: CVPR 2022. pp. 20730--20740 (2022)

\bibitem{touvron2021training}
Touvron, H., Cord, M., Douze, M., Massa, F., Sablayrolles, A., J{\'e}gou, H.:
  Training data-efficient image transformers \& distillation through attention.
  In: ICML 2021. pp. 10347--10357. PMLR (2021)

\bibitem{wang2021boundary}
Wang, J., Wei, L., Wang, L., et~al.: Boundary-aware transformers for skin
  lesion segmentation. In: MICCAI 2021. pp. 206--216. Springer (2021)

\bibitem{wu2022fat}
Wu, H., Chen, S., et~al.: {FAT-Net}: Feature adaptive transformers for
  automated skin lesion segmentation. Medical image analysis  \textbf{76},
  102327 (2022)

\bibitem{wu2023medical}
Wu, J., Fu, R., Fang, H., Liu, Y., Wang, Z., Xu, Y., Jin, Y., Arbel, T.:
  Medical {SAM} adapter: Adapting segment anything model for medical image
  segmentation. arXiv preprint arXiv:2304.12620  (2023)

\bibitem{xie2020mutual}
Xie, Y., Zhang, J., Xia, Y., Shen, C.: A mutual bootstrapping model for
  automated skin lesion segmentation and classification. IEEE transactions on
  medical imaging  \textbf{39}(7),  2482--2493 (2020)

\bibitem{yan2019melanoma}
Yan, Y., Kawahara, J., Hamarneh, G.: Melanoma recognition via visual attention.
  In: IPMI 2019. pp. 793--804. Springer (2019)

\bibitem{yang2023aim}
Yang, T., Zhu, Y., Xie, Y., Zhang, A., Chen, C., Li, M.: {AIM}: Adapting image
  models for efficient video action recognition. In: ICLR 2023 (2023)

\bibitem{zhang2021transfuse}
Zhang, Y., Liu, H., Hu, Q.: {TransFuse}: Fusing transformers and cnns for
  medical image segmentation. In: MICCAI 2021. pp. 14--24. Springer (2021)

\end{thebibliography}




% LNCS Homepage, \url{http://www.springer.com/lncs}. Last accessed 4
% \end{thebibliography}
\end{document}

\title{Supplementary Material: \modelname{}: Adapting Vision Transformers for Small Skin Lesion Segmentation Datasets}
\titlerunning{Supplementary Material: \modelname{}}
% \author{Paper ID: 14, Anonymous submission to ISICW 2023}
% %
% \authorrunning{Anonymous et al.}
\authorrunning{S. Du et al.}

\author{Siyi Du\inst{1}\orcidID{0000-0002-9961-4533} \and
Nourhan Bayasi\inst{1}\orcidID{0000-0003-4653-6081} \and
Ghassan Hamarneh\inst{2}\orcidID{0000-0001-5040-7448} \and
Rafeef Garbi\inst{1}\orcidID{0000-0001-6224-0876}}
% \institute{Anonymous Organization \\
% \email{abc@def.hij}}
\institute{University of British Columbia, Vancouver, British Columbia, CA \\
\email{\{siyi,nourhanb,rafeef\}@ece.ubc.ca} \\
\and Simon Fraser University, Burnaby, British Columbia, CA \\
\email{hamarneh@sfu.ca}} 
\maketitle              % typeset the header of the contribution

\begin{table*}
\centering
\caption{Skin lesion segmentation (SLS) results comparing BASE (\modelname{} w/o both adapters and the prompt generator and is fully fine-tuned), \modelname{}, and SOTA methods. We report the models' parameter count in millions (M). The 2nd column shows which pre-trained backbone the model used. R-34/50 represents ResNet-34/50.}
\label{table:SOTA}
\resizebox{\textwidth}{!}{
\begin{tabular}{|p{21mm}|P{21mm}|P{14mm}P{14mm}P{14mm}P{14mm}P{14mm}|P{14mm}P{14mm}P{14mm}P{14mm}P{14mm}|}
\hline 
\textbf{Model} & \textbf{Pre-}  & \multicolumn{10}{c|}{\textbf{Segmentation Results in Test Sets (\%)}} \\
\cline{3-12}
~ & \textbf{trained} & \multicolumn{5}{c|}{\textbf{Dice\tiny{\emph{$\pm$std}} \normalsize$\uparrow$}} & \multicolumn{5}{c|}{\textbf{IOU\tiny{\emph{$\pm$std}} \normalsize$\uparrow$}} \\
\cline{3-12}
~ & \textbf{backbone} &  ISIC & DMF & SCD & PH2 & Avg  & ISIC & DMF & SCD & PH2 & Avg\\
\hline 
\multicolumn{12}{|c|}{\textbf{(a) BASE \& Proposed Method}} \\ \hline
BASE	&	ViT-B	&	90.77\tiny{\emph{0.46}} 	&	91.69\tiny{\emph{0.79}} 	&	91.95\tiny{\emph{1.79}} 	&	95.64\tiny{\emph{0.73}} 	&	92.51\tiny{\emph{0.22}} 	&	83.71\tiny{\emph{0.70}} 	&	84.89\tiny{\emph{1.24}} 	&	85.42\tiny{\emph{2.82}} 	&	91.72\tiny{\emph{1.28}} 	&	86.43\tiny{\emph{0.34}} 	\\
\modelname{}	&	ViT-B	&	\underline{91.74}\tiny{\emph{0.64}} 	&	\underline{92.04}\tiny{\emph{0.67}} 	&	93.16\tiny{\emph{1.15}} 	&	95.66\tiny{\emph{0.56}} 	&	93.15\tiny{\emph{0.42}} 	&	\underline{85.22}\tiny{\emph{1.00}} 	&	\underline{85.47}\tiny{\emph{1.05}} 	&	87.39\tiny{\emph{1.94}} 	&	91.72\tiny{\emph{1.01}} 	&	87.45\tiny{\emph{0.70}} 	\\ \hline
\multicolumn{12}{|c|}{\textbf{(b) PEFT Methods}} \\ \hline
VPT	&	ViT-B	&	90.89\tiny{\emph{0.77}} 	&	91.26\tiny{\emph{0.58}} 	&	89.09\tiny{\emph{1.46}} 	&	93.14\tiny{\emph{0.85}} 	&	91.10\tiny{\emph{0.46}} 	&	83.83\tiny{\emph{1.17}} 	&	84.14\tiny{\emph{0.90}} 	&	80.76\tiny{\emph{2.23}} 	&	87.27\tiny{\emph{1.48}} 	&	84.00\tiny{\emph{0.74}} 	\\
AdaptFormer	&	ViT-B	&	91.12\tiny{\emph{0.74}} 	&	91.27\tiny{\emph{0.65}} 	&	89.65\tiny{\emph{0.75}} 	&	93.76\tiny{\emph{0.73}} 	&	91.45\tiny{\emph{0.42}} 	&	84.15\tiny{\emph{1.15}} 	&	84.18\tiny{\emph{1.00}} 	&	81.49\tiny{\emph{1.13}} 	&	88.33\tiny{\emph{1.30}} 	&	84.54\tiny{\emph{0.67}} 	\\ \hline

\multicolumn{12}{|c|}{\textbf{(c) SLS Methods w/o Pre-trained Backbones \& Trained From Scratch}} \\ \hline
SwinUnet	&	None	&	89.64\tiny{\emph{0.39}} 	&	90.67\tiny{\emph{0.73}} 	&	89.77\tiny{\emph{2.08}} 	&	94.24\tiny{\emph{0.93}} 	&	91.08\tiny{\emph{0.57}} 	&	81.94\tiny{\emph{0.62}} 	&	83.19\tiny{\emph{1.14}} 	&	82.07\tiny{\emph{3.02}}	&	89.24\tiny{\emph{1.59}} 	&	84.11\tiny{\emph{0.79}} 	\\
UNETR  &  None  & 89.60\tiny{\emph{0.61}} 	&	90.53\tiny{\emph{1.03}} 	&	88.13\tiny{\emph{3.67}} 	&	93.92\tiny{\emph{0.94}} 	&	90.55\tiny{\emph{0.87}} 	&	81.86\tiny{\emph{0.84}} 	&	83.02\tiny{\emph{1.60}} 	&	79.96\tiny{\emph{5.23}} 	&	88.68\tiny{\emph{1.58}} 	&	83.38\tiny{\emph{1.24}} 	  \\ 
UTNet	&	None	&	89.68\tiny{\emph{0.90}} 	&	89.87\tiny{\emph{0.58}} 	&	88.11\tiny{\emph{3.20}} 	&	93.29\tiny{\emph{1.08}} 	&	90.23\tiny{\emph{0.61}} 	&	81.99\tiny{\emph{1.25}} 	&	81.91\tiny{\emph{0.92}} 	&	79.71\tiny{\emph{4.34}} 	&	87.62\tiny{\emph{1.80}} 	&	82.81\tiny{\emph{0.77}} 	\\
MedFormer	&	None	&	90.47\tiny{\emph{0.68}} 	&	90.85\tiny{\emph{0.60}} 	&	90.60\tiny{\emph{2.56}} 	&	94.82\tiny{\emph{0.81}} 	&	91.68\tiny{\emph{0.74}} 	&	83.22\tiny{\emph{0.96}} 	&	83.52\tiny{\emph{0.95}} 	&	83.53\tiny{\emph{3.94}} 	&	90.23\tiny{\emph{1.44}} 	&	85.13\tiny{\emph{1.12}} 	\\
Swin UNETR	&	None	&	90.19\tiny{\emph{0.50}} 	&	91.00\tiny{\emph{0.72}} 	&	90.71\tiny{\emph{2.19}} 	&	94.54\tiny{\emph{0.77}} 	&	91.61\tiny{\emph{0.49}} 	&	82.78\tiny{\emph{0.76}} 	&	83.77\tiny{\emph{1.11}} 	&	83.54\tiny{\emph{3.45}} 	&	89.74\tiny{\emph{1.34}} 	&	84.96\tiny{\emph{0.74}} 	\\ \hline

\multicolumn{12}{|c|}{\textbf{(d) SLS Methods w/ Pre-trained Backbones \& Fully Fine-tuned}} \\ \hline
H2Former	&	R-34	&	91.17\tiny{\emph{0.72}} 	&	91.29\tiny{\emph{0.88}} 	&	92.76\tiny{\emph{1.81}} 	&	95.65\tiny{\emph{0.74}} 	&	92.72\tiny{\emph{0.63}} 	&	84.35\tiny{\emph{1.08}} 	&	84.22\tiny{\emph{1.38}} 	&	87.04\tiny{\emph{2.70}} 	&	91.77\tiny{\emph{1.30}} 	&	86.85\tiny{\emph{0.91}} 	\\
FAT-Net	&	R-34, DeiT-T	&	91.26\tiny{\emph{0.63}} 	&	91.32\tiny{\emph{0.73}} 	&	93.03\tiny{\emph{1.32}} 	&	96.07\tiny{\emph{0.52}} 	&	92.92\tiny{\emph{0.48}} 	&	84.42\tiny{\emph{0.98}} 	&	84.25\tiny{\emph{1.17}} 	&	87.23\tiny{\emph{2.19}} 	&	92.48\tiny{\emph{0.94}} 	&	87.10\tiny{\emph{0.80}} 	\\
BAT	&	R-50		&	91.33\tiny{\emph{0.68}} 	&	91.20\tiny{\emph{0.80}} 	&	92.95\tiny{\emph{1.40}} 	&	95.84\tiny{\emph{0.43}} 	&	92.83\tiny{\emph{0.46}} 	&	84.40\tiny{\emph{1.09}} 	&	84.03\tiny{\emph{1.29}} 	&	87.08\tiny{\emph{2.39}} 	&	92.04\tiny{\emph{0.78}} 	&	86.89\tiny{\emph{0.78}} 	\\
TransFuse	&	R-50, DeiT-B	&	91.73\tiny{\emph{0.51}} 	&	91.96\tiny{\emph{0.71}} 	&	\underline{94.11}\tiny{\emph{1.03}} 	&	\underline{96.18}\tiny{\emph{0.42}} 	&	\underline{93.50}\tiny{\emph{0.27}} 	&	\underline{85.22}\tiny{\emph{0.79}} 	&	85.33\tiny{\emph{1.15}} 	&	\underline{89.03}\tiny{\emph{1.75}} 	&	\underline{92.69}\tiny{\emph{0.76}} 	&	\underline{88.07}\tiny{\emph{0.47}} 	\\ \hline
\end{tabular}}
\end{table*}

\begin{table*}
\centering
\caption{Experiments using different pre-trained ViT backbones and ablation study of \modelname{}. $^*$ means the pre-trained backbone is frozen throughout training. $^{-P}$ or $^{-A}$ represent not using the prompt generator or adapters in \modelname{}.}
\label{table:ablation}
\resizebox{\textwidth}{!}{
\begin{tabular}{|p{15mm}|P{16mm}|P{14mm}P{14mm}P{14mm}P{14mm}P{14mm}|P{14mm}P{14mm}P{14mm}P{14mm}P{14mm}|}
\hline 
\textbf{Model} &  \textbf{Pre-} & \multicolumn{10}{c|}{\textbf{Segmentation Results in Test Sets (\%)}} \\
\cline{3-12}
~ & \textbf{trained} & \multicolumn{5}{c|}{\textbf{Dice\tiny{\emph{$\pm$std}} \normalsize$\uparrow$}} & \multicolumn{5}{c|}{\textbf{IOU\tiny{\emph{$\pm$std}} \normalsize$\uparrow$}} \\
\cline{3-12}
~ & \textbf{backbone} &  ISIC & DMF & SCD & PH2 & Avg  & ISIC & DMF & SCD & PH2 & Avg\\
\hline
\multicolumn{12}{|c|}{\textbf{(a) Applicability to Various Pre-trained ViT Backbones}} \\ \hline
BASE	&	Swin-B	&	91.63\tiny{\emph{0.71}} 	&	91.70\tiny{\emph{0.62}} 	&	92.71\tiny{\emph{0.99}} 	&	95.88\tiny{\emph{0.34}} 	&	92.98\tiny{\emph{0.37}} 	&	85.05\tiny{\emph{1.11}} 	&	84.89\tiny{\emph{0.99}} 	&	86.60\tiny{\emph{1.61}} 	&	92.13\tiny{\emph{0.61}} 	&	87.17\tiny{\emph{0.61}}  	\\
\modelname{}	&	Swin-B	&	91.54\tiny{\emph{0.60}} 	&	91.73\tiny{\emph{0.71}} 	&	93.60\tiny{\emph{0.95}} 	&	95.68\tiny{\emph{0.70}} 	&	93.14\tiny{\emph{0.39}}  	&	84.90\tiny{\emph{0.96}} 	&	84.94\tiny{\emph{1.12}} 	&	88.12\tiny{\emph{1.62}} 	&	91.77\tiny{\emph{1.26}} 	&	87.43\tiny{\emph{0.64}}  	\\
BASE	&	Swin-L	&	91.64\tiny{\emph{0.65}} 	&	91.69\tiny{\emph{0.65}} 	&	92.93\tiny{\emph{1.03}} 	&	95.83\tiny{\emph{0.46}} 	&	93.02\tiny{\emph{0.25}}  	&	85.08\tiny{\emph{1.05}} 	&	84.86\tiny{\emph{1.04}} 	&	86.97\tiny{\emph{1.66}} 	&	92.04\tiny{\emph{0.82}} 	&	87.24\tiny{\emph{0.43}} 	\\
\modelname{}	&	Swin-L	&	91.56\tiny{\emph{0.63}} 	&	91.91\tiny{\emph{0.56}} 	&	93.74\tiny{\emph{0.98}} 	&	96.07\tiny{\emph{0.50}} 	&	93.32\tiny{\emph{0.31}}  	&	84.93\tiny{\emph{1.00}} 	&	85.24\tiny{\emph{0.87}} 	&	88.38\tiny{\emph{1.69}} 	&	92.47\tiny{\emph{0.92}} 	&	87.76\tiny{\emph{0.50}}  	\\
BASE	&	ViT-L	&	91.37\tiny{\emph{0.78}} 	&	91.76\tiny{\emph{0.81}} 	&	93.23\tiny{\emph{1.04}} 	&	95.86\tiny{\emph{0.59}} 	&	93.06\tiny{\emph{0.29}}  	&	84.60\tiny{\emph{1.18}} 	&	84.99\tiny{\emph{1.29}} 	&	87.52\tiny{\emph{1.65}} 	&	92.09\tiny{\emph{1.06}} 	&	87.30\tiny{\emph{0.47}}  	\\
\modelname{}	&	ViT-L	&	91.54\tiny{\emph{0.59}} 	&	91.77\tiny{\emph{0.71}} 	&	93.48\tiny{\emph{1.16}} 	&	95.73\tiny{\emph{0.60}} 	&	93.13\tiny{\emph{0.48}}  	&	84.88\tiny{\emph{0.95}} 	&	85.01\tiny{\emph{1.11}} 	&	87.94\tiny{\emph{1.94}} 	&	91.85\tiny{\emph{1.08}} 	&	87.42\tiny{\emph{0.79}}  	\\  
BASE	&	DeiT-B	&		91.48\tiny{\emph{0.72}} 	&	91.82\tiny{\emph{0.85}} 	&	93.63\tiny{\emph{1.23}} 	&	95.83\tiny{\emph{0.45}} 	&	92.94\tiny{\emph{0.32}}  	&	84.77\tiny{\emph{1.10}}	&	85.10\tiny{\emph{1.36}} 	&	86.53\tiny{\emph{1.92}} 	&	92.04\tiny{\emph{0.81}} 	&	87.11\tiny{\emph{0.52}}  	\\
\modelname{}	&	DeiT-B	&		91.70\tiny{\emph{0.65}} 	&	91.85\tiny{\emph{0.82}} 	&	93.67\tiny{\emph{0.88}} 	&	95.97\tiny{\emph{0.46}} 	&	93.30\tiny{\emph{0.31}}  	&	85.14\tiny{\emph{1.05}} 	&	85.17\tiny{\emph{1.28}} 	&	88.22\tiny{\emph{1.51}} 	&	92.30\tiny{\emph{0.84}} 	&	87.71\tiny{\emph{0.51}}  	\\  
\hline

\multicolumn{12}{|c|}{\textbf{(b) Ablation Study}} \\ \hline
BASE$^*$	&	ViT-B	
&	87.18\tiny{\emph{0.84}} 	&	89.23\tiny{\emph{0.88}} 	&	86.24\tiny{\emph{1.55}} 	&	90.17\tiny{\emph{1.18}} 	&	88.20\tiny{\emph{0.46}}  	&	77.92\tiny{\emph{1.14}} 	&	80.81\tiny{\emph{1.31}} 	&	76.27\tiny{\emph{2.25}} 	&	82.30\tiny{\emph{1.87}} 	&	79.33\tiny{\emph{0.65}}  	\\
\modelname{}$^{-P}$	&	ViT-B	
&	91.47\tiny{\emph{0.63}} 	&	91.80\tiny{\emph{0.58}} 	&	91.18\tiny{\emph{0.79}} 	&	94.75\tiny{\emph{0.65}} 	&	92.30\tiny{\emph{0.31}}  	&	84.74\tiny{\emph{0.96}} 	&	85.04\tiny{\emph{0.90}} 	&	83.98\tiny{\emph{1.26}} 	&	90.09\tiny{\emph{1.16}} 	&	85.96\tiny{\emph{0.48}}  	\\
\modelname{}$^{-A}$	&	ViT-B	
&	90.87\tiny{\emph{0.72}} 	&	91.00\tiny{\emph{0.68}} 	&	89.09\tiny{\emph{3.62}} 	&	93.87\tiny{\emph{0.64}} 	&	91.21\tiny{\emph{0.83}}  	&	83.78\tiny{\emph{1.10}} 	&	83.72\tiny{\emph{1.06}} 	&	81.18\tiny{\emph{5.16}} 	&	88.53\tiny{\emph{1.14}} 	&	84.30\tiny{\emph{1.19}}  	\\
\modelname{}	&	ViT-B	&	\underline{91.74}\tiny{\emph{0.64}} 	&	\underline{92.04}\tiny{\emph{0.67}} 	&	\underline{93.16}\tiny{\emph{1.15}} 	&	\underline{95.66}\tiny{\emph{0.56}} 	&	\underline{93.15}\tiny{\emph{0.42}} 	&	\underline{85.22}\tiny{\emph{1.00}} 	&	\underline{85.47}\tiny{\emph{1.05}} 	&	\underline{87.39}\tiny{\emph{1.94}} 	&	\underline{91.72}\tiny{\emph{1.01}} 	&	\underline{87.45}\tiny{\emph{0.70}} 	\\ \hline
\end{tabular}}
\end{table*}